\documentclass[runningheads]{llncs}
\usepackage{graphicx}
\usepackage{comment}
\usepackage{amsmath}
\usepackage{amsfonts}
\usepackage{url}
\usepackage{multirow}
\usepackage{enumitem}
\usepackage{algpseudocode}
\usepackage{algorithm}

\usepackage{xcolor}
\newtheorem{prop}{Proposition}

\usepackage[title]{appendix}
\usepackage{multicol}
\usepackage{rotating}

\begin{document}
\title{GanDef: A GAN based Adversarial Training Defense for Neural Network Classifier}
\titlerunning{GanDef}
%
\author{Guanxiong Liu\inst{1} \and
Issa Khalil\inst{2} \and
Abdallah Khreishah\inst{1}}
\authorrunning{Guanxiong Liu et al.}
%
\institute{New Jersey Institute of Technology, Newark NJ 07102, USA \and
Qatar Computing Research Institute, Doha, Qatar \\
\email{gl236@njit.edu}, \email{ikhalil@hbku.edu.qa}, \email{abdallah@njit.edu}}
\maketitle              
\begin{abstract}
Machine learning models, especially neural network (NN) classifiers, are widely used in many applications including natural language processing, computer vision and cybersecurity. They provide high accuracy under the assumption of attack-free scenarios. However, this assumption has been defied by the introduction of adversarial examples -- carefully perturbed samples of input that are usually misclassified. Many researchers have tried to develop a defense against adversarial examples; however, we are still far from achieving that goal. In this paper, we design a Generative Adversarial Net (GAN) based adversarial training defense, dubbed \textbf{GanDef}, which utilizes a competition game to regulate the feature selection during the training. We analytically show that GanDef can train a classifier so it can defend against adversarial examples. Through extensive evaluation on different white-box adversarial examples, the classifier trained by GanDef shows the same level of test accuracy as those trained by state-of-the-art adversarial training defenses. More importantly, \textbf{GanDef-Comb}, a variant of GanDef, could utilize the discriminator to achieve a dynamic trade-off between correctly classifying original and adversarial examples. As a result, it achieves the highest overall test accuracy when the ratio of adversarial examples exceeds 41.7\%.
\keywords{Neural Network Classifier, Generative Adversarial Net, Adversarial Training Defense.}
\end{abstract}

\section{Introduction} \label{sec:intro}

Due to the surprisingly good representation power of complex distributions, NN models are widely used in many applications including natural language processing, computer vision and cybersecurity. For example, in cybersecurity, NN based classifiers are used for spam filtering, phishing detection as well as face recognition \cite{rowley1998neural} \cite{abu2007comparison}. However, the training and usage of NN classifiers are based on an underlying assumption that the environment is attack free. Therefore, such classifiers fail when adversarial examples are presented to them. 

Adversarial examples were first introduced in \cite{szegedy2013intriguing} in the context of image classification. It shows that a visually insignificant modification with specially designed perturbations can result in a huge change of prediction results with nearly $100\%$ success rate. Generally, adversarial examples can be used to mislead NN models to output any aimed prediction. They could be extremely harmful for many applications that utilize NNs, such as automatic cheque withdrawal in banks, traffic speed detection, and medical diagnosis in hospitals. As a result, this serious threat inspires a new line of research to explore the vulnerability of NN classifiers and develop appropriate defensive methods.

Recently, a plethora of methods to countermeasure adversarial examples has been introduced and evaluated. Among these methods, adversarial training defenses play an important role since they (1) effectively enhance the robustness, and (2) do not limit adversary's knowledge. However, most of them lack the trade-off between classifying original and adversarial examples. For applications that are sensitive to misbehavior or operate in risky environment, it is worth to enhance defenses against adversarial examples by sacrificing performance on original examples. The ability to dynamically control such trade-off makes the defense even more valuable.

In this paper, we propose a GAN based defense against adversarial examples, dubbed \textbf{GanDef}. GanDef is designed based on adversarial training combined with feature learning \cite{louppe2017learning}~\cite{xie2017controllable}~\cite{lample2017fader}. As a GAN model, GanDef contains a classifier and a discriminator which form a minimax game. To achieve the dynamic trade-off between classifying original and adversarial examples, we also propose a variant of GanDef, \textbf{GanDef-Comb}, that utilizes both classifier and discriminator. During evaluation, we select several state-of-the-art adversarial training defenses as references, including Pure PGD training (\textbf{Pure PGD}) \cite{madry2017towards}, Mix PGD training (\textbf{Mix PGD}) \cite{kannan2018adversarial} and \textbf{Logit Pairing} \cite{kannan2018adversarial}. The comparison results show that GanDef performs better than state-of-the-art adversarial training defenses in terms of test accuracy. Our contributions can be summarized as follows:
\begin{itemize}
    \item We propose the defensive method, GanDef, which is based on the idea of using a discriminator to regularize classifier's feature selection.
    
    \item We mathematically prove that the solution of the proposed minimax game in GanDef contains an optimal classifier, which usually makes correct predictions on adversarial examples by using perturbation invariant features.
    
    \item We empirically show that the trained classifier in GanDef achieves the same level of test accuracy as that in state-of-the-art approaches. Adding the discriminator, GanDef-Comb can dynamically control the trade-off on classifying original and adversarial examples and achieves the highest overall test accuracy when the ratio of adversarial examples exceeds 41.7\%.
\end{itemize}

The rest of the paper is organized as follows: Section \ref{sec:background} presents background material, Section \ref{sec:defense} details the design and mathematical proof of GanDef, Section \ref{sec:results} shows evaluation results, Section \ref{sec:conclusion} concludes the paper, and Section \ref{sec:future} discusses the future work.

\section{Background and Related Work} \label{sec:background}

In this section, we introduce high-level background material about threat model, adversarial example generators and defensive mechanisms for the better understanding of concepts presented in this work. We also provide relevant references for further information about each topic.

\subsection{Threat Model}

The adversary aims at misleading the NN model utilized by the application to achieve a malicious goal. For example, adversary adds adversarial perturbation to the image of a cheque. As a result, this image may mislead the NN model utilized by the ATM machine to cash out a huge amount of money. During the preparation of adversarial examples we assume that adversary has full knowledge of the targeted NN model, which is the white-box scenario. Also, we assume that adversary has limited computational power. As a result, the adversary can generate iterative adversarial examples but cannot exhaustively search all possible input perturbation.

\subsection{Generating Adversarial Examples}

The adversarial examples could be classified into white-box and black-box attacks based on adversary's knowledge of target NN classifier. Based on the generating process, they could be also classified as single-step and iterative adversarial examples.

\textbf{Fast Gradient Sign Method (FGSM)} is introduced by Goodfellow et. al in \cite{goodfellow2014explaining} as a single-step white-box adversarial example generator against NN image classifiers. This method tries to maximize the loss function value, $\mathcal{L}$, of NN classifier, $\mathcal{C}$, to find adversarial examples. The function $\mathcal{F}$ is used to ensure that the generated adversarial example is still a valid image.
\begin{equation*}
\begin{aligned}
    & \underset{\delta}{\text{maximize}}
    & & \mathcal{L} (\hat{z}=\mathcal{C}(\hat{x}), t)
    & \text{subject to}
    & & \hat{x} = \mathcal{F} (\bar{x}, \delta) \in \mathbb{R}_{[0,1]}^{m}
\end{aligned}
\end{equation*}
To keep visual similarity and enhance generation speed, this maximization problem is solved by running gradient ascent for one iteration. It simply generates adversarial examples, $\hat{x}$, from original images, $\bar{x}$, by adding small perturbation, $\delta$, which changes each pixel value along the gradient direction of the loss function. As a single step adversarial example generator, FGSM can generate adversarial examples efficiently. However, the quality of the generated adversarial examples is relatively low due to the linear approximation of the loss function landscape.

\textbf{Basic Iterative Method (BIM)} is introduced by Kurakin et. al in \cite{kurakin2016adversarial} as an iterative white-box adversarial example generator against NN image classifiers. In the algorithm design, BIM utilizes the same mathematical model as FGSM. But, different from the FGSM, BIM is an iterative attack method. Instead of making the adversarial perturbation in one iteration, BIM runs the gradient ascent algorithm multiple iterations to maximize the loss function. In each iteration, BIM applies smaller perturbation and maps the perturbed image through the function $\mathcal{F}$. As a result, BIM approximates the loss function landscape by linear spline interpolation. Therefore, it generates stronger adversarial examples than FGSM within the same neighboring area.

\textbf{Projected Gradient Descent (PGD)} is another iterative white-box adversarial example generator recently introduced by Madry et. al in \cite{madry2017towards}. Similar to BIM, PGD also solves the same optimization problem iteratively with the projected gradient descent algorithm. However, PGD randomly selects an initial point within a limited area of the original image and repeats this several times to generate an adversarial example. With this multiple time random initialization, PGD is shown experimentally to solve the optimization problem efficiently and generate more serious adversarial examples since the loss landscape has a surprisingly tractable structure \cite{madry2017towards}.

\subsection{Adversarial Example Defensive Methods}

Many defense methods have been proposed recently. In the following, we summarize and present representative samples from three major defense classes.

\textbf{Augmentation and Regularization} aims at penalizing overconfident prediction or utilizing synthetic data during training. One of the early ideas is the defensive distillation. Defensive distillation uses the prediction score from original NN, usually called teacher, as ground truth to train another smaller NN, usually called student \cite{papernot2016distillation} \cite{papernot2017extending}. It has been shown that the calculated gradients from the student model become very small or even reach zero and hence become useless to the adversarial example generator \cite{papernot2017extending}. Some of the recent works that belong to this set of methods are referred to as Fortified Network \cite{lamb2018fortified} and Manifold Mixup \cite{verma2018manifold}. Fortified Network utilizes denoising autoencoder to regularize the hidden states. Manifold Mixup also focuses on the hidden states but follows a different way. During the training, Manifold Mixup uses interpolations of hidden states and logits during training to enhance the diversity of training data. Compared with adversarial training defenses, this set of defenses has significant limitations. For example, defensive distillation is vulnerable to Carlini attack \cite{carlini2017towards} and Manifold Mixup can only defend against single step attacks.

\textbf{Protective Shell} is a set of defensive methods which aim at using a shell to reject or reform the adversarial examples. An example of these methods is introduced by Meng et. al in \cite{meng2017magnet} which is called MagNet. In this work, the authors design two types of functional components: the detector and the reformer. Adversarial examples are either rejected by the detector or reformed to eliminate the perturbations. Other recent works such as \cite{liang2017detecting} and \cite{samangouei2018defense} try to utilize different methods to build the shell. In \cite{liang2017detecting}, authors inject adaptive noise to input images which breaks the adversarial perturbations without significant decrease of classification accuracy. In \cite{samangouei2018defense}, a generator is utilized to generate images that are similar to the inputs. By replacing the inputs with generated images, it achieves resistance to adversarial examples. However, this set of methods usually assume the shell itself is black-box to the adversary and the work in \cite{athalye2018obfuscated} has already found ways to break such an assumption.

\textbf{Adversarial Training} is based on a straightforward idea that treats adversarial examples as blind spots of the original training data \cite{xu2016automatically}. Through retraining with adversarial examples, the classifier learns the perturbation pattern and generalizes its prediction to account for such perturbations. In \cite{goodfellow2014explaining}, the adversarial examples generated by FGSM are used for adversarial training and the trained NN classifier can defend single step adversarial examples. Later works in \cite{madry2017towards} and \cite{tramer2017ensemble} enhance the adversarial training method to defend examples like BIM and PGD. A more recent work in \cite{kannan2018adversarial} requires that the pre-softmax logits from original and adversarial examples to be similar. Authors believe this method could utilize more information during adversarial training. A common problem in existing adversarial training defenses is that the trained classifier has no control of the trade-off between correctly classifying original and adversarial examples. Our work achieves this flexibility and shows the benefit.

\section{GanDef: GAN based Adversarial Training} \label{sec:defense}

In this section, we present the design of our defensive method (GanDef) as follows. First, the design of GanDef is introduced as a minimax game of the classifier and discriminator. Then we conduct a theoretical analysis of the proposed minimax game in GanDef. Finally, we conduct experimental analysis to evaluate the convergence of GanDef.

\subsection{Design}

Given the training data pair $\langle x, t \rangle$, where $x \in \cup (\bar{X}, \hat{X})$, we try to find a classification function $\mathcal{C}$ that uses $x$ to produce pre-softmax logits $z$ such that:
\begin{equation*}
\begin{aligned}
    & t_{i} = f(z_{i}) = \frac{e^{z_{i}}}{\sum_{z_{j}} e^{z_{j}}} & & \text{The mapping between $z$ and $t$ is the softmax function.}
\end{aligned}
\end{equation*}
Since $x$ can be either original example $\bar{x}$ or adversarial example $\hat{x}$, we want the classifier to model the conditional probability $q_{C}(z|x)$ with only non-adversarial features. To achieve this, we employ another NN and call it discriminator $\mathcal{D}$. $\mathcal{D}$ uses the pre-softmax logits $z$ from $\mathcal{C}$ as inputs and predicts whether the input to classifier is $\bar{x}$ or $\hat{x}$. This process can be performed by maximizing the conditional probability $q_{D}(s|z)$, where $s$ is a Boolean variable indicating the source of $x$ is original or adversarial. Finally, by combining the classifier and the discriminator, we formulate the following minimax game:
\begin{equation*}
\begin{aligned}
    & \underset{\mathcal{C}}{\text{min}} ~ \underset{\mathcal{D}}{\text{max}} ~ J(\mathcal{C}, \mathcal{D})
\end{aligned}
\end{equation*}
\begin{equation*}
\begin{aligned}
    & \text{where} ~~ J(\mathcal{C}, \mathcal{D}) = 
    & \underset{x \sim X, t \sim T}{\mathbb{E}} \{- log [q_{C}(z|x)]\}  - \underset{z \sim Z, s \sim S}{\mathbb{E}} \{- log [q_{D}(s|z=\mathcal{C}(x))]\}
\end{aligned}
\end{equation*}

In this work, we envision that the classifier could be seen as a generator that generates pre-softmax logits based on selected features from input images. Then, the classifier and the discriminator engage in a minimax game, which is also known as \textit{Generative Adversarial Net} (GAN) \cite{goodfellow2016deep}. Therefore, we name our proposed defense as ``GAN based Adversarial Training'' (GanDef). While other defenses ignore or only compare $\bar{z}$ and $\hat{z}$, utilizing discriminator with $z$ adds a second line of defense when the classifier is defeated by adversarial examples.

\begin{figure*}[tb]
\centering
\begin{minipage}[c]{.45\textwidth}
\centering
    \includegraphics[width=\linewidth]{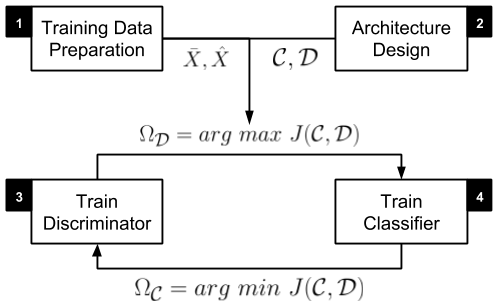}
    \caption{Training with GanDef}
    \label{fig:gan-example}
\end{minipage}
\begin{minipage}[c]{.45\textwidth}
\centering
    \includegraphics[width=\linewidth]{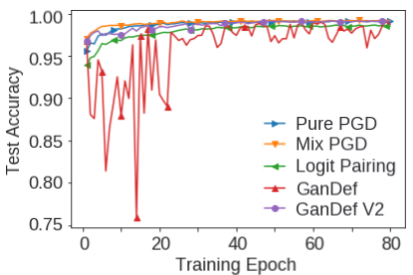}
    \caption{Convergence Experiments}
    \label{fig:convergence}
\end{minipage}
\vspace{-5mm}
\end{figure*}

The pseudocode of GanDef training is summarized in Algorithm \ref{algorithm:defense} and is visualized in Figure \ref{fig:gan-example}. A summary of the notations used throughout this work is available in Table \ref{table:summary-notation}.
\begin{table*}[tb]
    \begin{center}
    \begin{tabular}{ p{.23\textwidth} | p{.77\textwidth} }
    \hline \hline
    $\mathcal{L}$
    & loss function of NN classifier \\
    $\mathcal{F}$
    & function which regularize pixel value of generated example \\
    $\bar{x}, ~ \bar{X}; ~ \hat{x}, ~ \hat{X}; ~ x, ~ X$ 
    & original, adversarial and all training examples \\
    $\bar{t}, ~ \bar{T}; ~ \hat{t}, ~ \hat{T}; ~ t, ~ T$ 
    & ground truth of original, adversarial and all training examples \\
    $\bar{z}, ~ \bar{Z}; ~ \hat{z}, ~ \hat{Z}; ~ z, ~ Z$ 
    & pre-softmax logits of original, adversarial and all training examples \\
    $\bar{s}, ~ \bar{S}; ~ \hat{s}, ~ \hat{S}; ~ s, ~ S$ 
    & source indicator of original, adversarial and all training examples \\
    $\delta$
    & adversarial perturbation \\
    $\mathcal{C}, ~ \mathcal{C}^{*}$
    & NN based classifier \\
    $\mathcal{D}, ~ \mathcal{D}^{*}$
    & NN based discriminator \\
    $J, ~ J'$
    & reward function of the minimax game \\
    $\Omega, ~ \Omega_{\mathcal{C}}, ~ \Omega_{\mathcal{D}}$
    & weight parameter in the NN model \\
    $\gamma$
    & trade-off hyper-parameters in GanDef \\
    \hline \hline
    \end{tabular}
    \end{center}
    \caption{Summary of Notations}
    \label{table:summary-notation}
\end{table*}
\begin{algorithm} \caption{GanDef Training} \label{algorithm:defense}
\begin{algorithmic}[1]
\Require training examples $X$, ground truth $T$, classifier $\mathcal{C}$, discriminator $\mathcal{D}$
\Ensure classifier $\mathcal{C}$, discriminator $\mathcal{D}$
\State Initialize weight parameters $\Omega$ in both classifier and discriminator
\For {the global training iterations}
    \For {the discriminator training iterations}
        \State Randomly sample a batch of training examples, $\langle x, t \rangle$ \label{alg2:sample1}
        \State Generate a batch of boolean indicator, $s$, corresponding to training inputs \label{alg2:prepare}
        \State Fix weight parameters $\Omega_{\mathcal{C}}$ in classifier $\mathcal{C}$ \label{alg2:freeze1}
        \State Update weight parameters $\Omega_{\mathcal{D}}$ in discriminator $\mathcal{D}$ by stochastic gradient descent \label{alg2:update1}
    \EndFor
    \State Randomly sample a batch of training examples, $\langle x, t \rangle$ \label{alg2:sample2}
    \State Generate a batch of boolean indicator, $s$, corresponding to training inputs
    \State Fix weight parameters $\Omega_{\mathcal{D}}$ in discriminator $\mathcal{D}$ \label{alg2:freeze2}
    \State Update weight parameters $\Omega_{\mathcal{C}}$ in classifier $\mathcal{C}$ by stochastic gradient descent \label{alg2:update2}
\EndFor
\end{algorithmic}
\end{algorithm}

\subsection{Theoretical Analysis}
With the formal definition of our GanDef, we perform a theoretical analysis in this subsection. We show that under the current definition where $J$ is a combination of log likelihood of $Z|X$ and $S|Z$, the solution of the minimax game contains an optimal classifier which can correctly classify adversarial examples. It is worth noting that our analysis is conducted in a non-parametric setting, which means that the classifier and the discriminator have enough capacity to model any distribution.

\begin{prop} \label{prop:1}
    If there exists a solution $(\mathcal{C}^{*}, \mathcal{D}^{*})$ for the aforementioned minmax game $J$ such that $J(\mathcal{C}^{*}, \mathcal{D}^{*}) = H(Z|X) - H(S)$, then $\mathcal{C}^{*}$ is a classifier that can defend against adversarial examples.
\end{prop}

\begin{proof} For any fixed classification model $\mathcal{C}$, the optimal discriminator can be formulated as
\begin{equation*}
\begin{aligned}
    \mathcal{D}^{*} &= \text{arg} ~ \underset{\mathcal{D}}{\text{max}} ~ J(\mathcal{C}, \mathcal{D})
    &= \text{arg} ~ \underset{\mathcal{D}}{\text{min}} ~ \underset{z \sim Z, s \sim S}{\mathbb{E}} \{- log [q_{D}(s|z=\mathcal{C}(x))]\}
\end{aligned}
\end{equation*}
In this case, the discriminator can perfectly model the conditional distribution and we have $q_{D}(s|z=\mathcal{C}(x)) = p(s|z=\mathcal{C}(x))$ for all $z$ and all $s$. Therefore, we can rewrite $J$ with optimal discriminator as $J'$ and denote the second half of $J$ as a conditional entropy $H(S|Z)$
\begin{equation*}
\begin{aligned}
    & J'(\mathcal{C}) = \underset{x \sim X, t \sim T}{\mathbb{E}} \{- log [q_{C}(z|x)]\} - H(S|Z)
\end{aligned}
\end{equation*}
For the optimal classification model, the goal is to achieve the conditional probability $q_{C}(z|x) = p(z|x)$ since $z$ can determine $t$ by taking softmax transformation. Therefore, the first part of $J'(\mathcal{C})$ (the expectation) is larger than or equal to $H(Z|X)$. Combined with the basic property of conditional entropy that $H(S|Z) \leq H(S)$, we can get the following lower bound of $J$ with optimal classifier and discriminator
\begin{equation*}
\begin{aligned}
    & J(\mathcal{C}^{*}, \mathcal{D}^{*}) \geq H(Z|X) - H(S|Z) \geq H(Z|X) - H(S)
\end{aligned}
\end{equation*}
This equality holds when the following two conditions are satisfied:
\begin{itemize}
    \item The classifier perfectly models the conditional distribution of $z$ given $x$, $q_{C}(z|x) = p(z|x)$, which means that $\mathcal{C}^{*}$ is an optimal classifier.
    \item $S$ and $Z$ are independent, $H(S|Z) = H(S)$, which means that adversarial perturbations do not affect pre-softmax logits.
\end{itemize}
\end{proof}

In practice, the assumption of unlimited capacity in classifier and discriminator may not hold and it would be hard or even impossible to build an optimal classifier that outputs pre-softmax logits that are independent from adversarial perturbation. Therefore, we introduce a trade-off hyper-parameter $\gamma$ into the minimax function as follows:
\begin{equation*}
\begin{aligned}
    & \underset{x \sim X, t \sim T}{\mathbb{E}} \{- log [q_{C}(z|x)]\}  - \gamma \underset{z \sim Z, s \sim S}{\mathbb{E}} \{- log [q_{D}(s|z=\mathcal{C}(x))]\}
\end{aligned}
\end{equation*}
When $\gamma = 0$, GanDef is the same as traditional adversarial training. When $\gamma$ increases, the discriminator becomes more and more sensitive to information of $s$ contained in pre-softmax logits, $z$.

\subsection{Convergence Analysis}

Beyond the theoretical analysis, we also conduct an experimental analysis of the convergence of GanDef. Based on the pseudocode in Algorithm \ref{algorithm:defense}, we train a classifier on MNIST dataset. In order to compare the convergence, we also implement Pure PGD, Mix PGD and Logit Pairing and present their test accuracies on original test images during different training epochs.

As we can see from Figure \ref{fig:convergence}, the convergence of GanDef is not as good as other state-of-the-art adversarial training defenses. Although all these methods converge to over 95\% test accuracy, GanDef shows significant fluctuation during the training process.

In order to improve the convergence of GanDef, we carefully trace back the design process and identify the root cause of the fluctuations. During the training of the classifier, we subtract the penalty term $\underset{z \sim Z, s \sim S}{\mathbb{E}} \{- log [q_{D}(s|z=\mathcal{C}(x))]\}$ which encourages the classifier to hide information of $s$ in every $z$. Compared with Logit Pairing which requires similar $z$ from original and adversarial examples, our penalty term is too strong. Therefore, we modify the training loss of the classifier to:
\begin{equation*}
\begin{aligned}
    & \underset{x \sim X, t \sim T}{\mathbb{E}} \{- log [q_{C}(z|x)]\}  - \gamma \underset{\hat{z} \sim \hat{Z}, \hat{s} \sim \hat{S}}{\mathbb{E}} \{- log [q_{D}(\hat{s}|\hat{z}=\mathcal{C}(\hat{x}))]\}
\end{aligned}
\end{equation*}
Recall that $\hat{x}$, $\hat{z}$ and $\hat{s}$ represent the adversarial example, its pre-softmax logits, and the source indicator, respectively. It is also worth to mention that this modification is only applied to the classifier. Therefore, it does not affect the consistency of the previous proof. During convergence analysis, we denote the modified version of our defensive method as GanDef V2 and its convergence results are also shown in Figure \ref{fig:convergence}. It is clear that GanDef V2 significantly improves the convergence and stability during the training. Moreover, its test accuracy on the original as well as several different white-box adversarial examples is also higher than the initial design. Due to these improvements, we use it as the standard implementation of GanDef in the rest of this work.

\section{Experiments and Results} \label{sec:results}

In this section, we present comparative evaluation results of the adversarial training defenses introduced previously.

\subsection{Datasets, NN Structures and Hyper-parameter}

During evaluation, we conduct experiments for classifying original and adversarial examples on both MNIST and CIFAR10 datasets. To ensure the quality of evaluation, we utilize the standard python library (CleverHans \cite{papernot2018cleverhans}) and run all experiments on a Linux Workstation with NVIDIA GTX-1080 GPU. We choose the adversarial examples introduced in Section \ref{sec:background} and denote them as FGSM, BIM, PGD-1 and PGD-2 examples. For MNIST dataset, PGD-1 represents 40-iteration PGD attack while PGD-2 corresponds to 80-iteration PGD attack. Moreover, the maximum perturbation limitation is 0.3. The per step perturbation limitations for BIM and PGD examples are 0.05 and 0.01. For CIFAR10 dataset, these two sets of adversarial examples are 7-iteration and 20-iteration PGD attack. The maximum perturbation limitation is $\frac{8}{255}$ while per step perturbation limitation for BIM and PGD is $\frac{2}{255}$.

During the training, the vanilla classifier only uses original training data while defensive methods utilize original and PGD-1 examples except for Pure PGD which only requires the PGD-1 examples. For the testing part, we generate adversarial examples based on test data which was not used in training. These adversarial examples together with original test data form the complete test dataset during the evaluation stage. To make a fair comparison, defensive methods and vanilla classifier share the same NN structures which are (1) LeNet \cite{madry2017towards} for MNIST, and (2) allCNN \cite{springenberg2014striving} for CIFAR10. Due to the page limitation, the detailed structure is shown in the Appendix. The hyper-parameter of existing defensive methods are the same as the original papers \cite{madry2017towards}\cite{kannan2018adversarial}. During the training of Logit Pairing on CIFAR10, we found that using the same trade-off parameter as MNIST lead to divergence. To resolve the issue, we try to change the optimizer, learning rate, initialization and weight decay. However, none of them work until the weight of logit comparison loss is decreased to 0.01.

To validate the NN structure as well as the adversarial examples, we utilize the vanilla classifier to classify original and adversarial examples. Based on the results in Table \ref{table:summary-TA}, the test accuracy of the vanilla classifier on original examples matches the records of benchmarks in \cite{benchmark-list}. Moreover, the test accuracy of the vanilla classifier on any kind of adversarial examples has significant degeneration which shows the adversarial example generators are working properly.

\subsection{Comparative Evaluation of Defensive Approaches}

\begin{figure*}[tb]
\centering
\begin{minipage}[c]{.9\textwidth}
\centering
    \includegraphics[width=\linewidth]{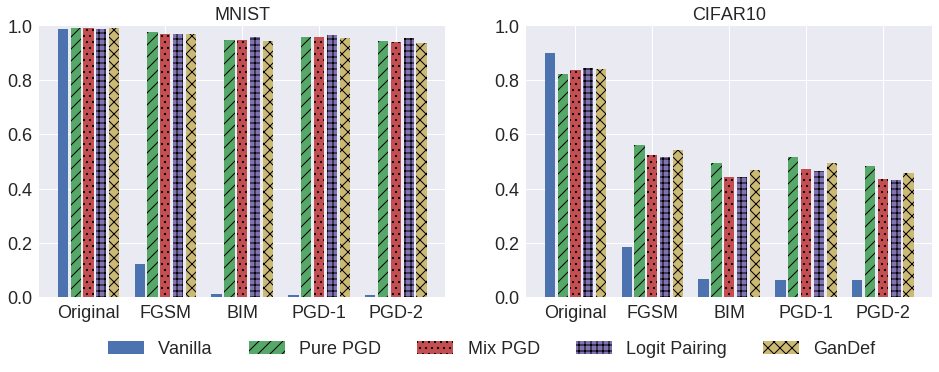}
\end{minipage}
\vspace{-4mm}
\caption{Visualization of Test Accuracy on Original and Adversarial Examples}
\label{fig:visualization-TA}
\end{figure*}

As the first step, we compare the GanDef with state-of-the-art adversarial training defenses in terms of test accuracy on original and white-box adversarial examples. The results are presented in Figure \ref{fig:visualization-TA} and summarized in Table \ref{table:summary-TA}.

On MNIST, all defensive methods achieve around 99\% test accuracy on original examples and Pure PGD is slightly better than others. In general, the test accuracy of defensive methods are almost the same and does not go lower than that of the vanilla model. On CIFAR10, we can see that the test accuracy of defensive methods on original data is around 83\% and these of Logit Pairing and GanDef are slightly higher than others. Compared with the vanilla classifier, there is about 5\% decrease in test accuracy. Similar degeneration is also reported in previous works on Pure PGD, Mix PGD and Logit Pairing \cite{madry2017towards}\cite{kannan2018adversarial}.

During the evaluation on MNIST, there are no significant differences among defensive methods and each could achieve around 95\% test accuracy. The Pure PGD method is the best on the evaluation of FGSM and BIM examples, while the Logit Pairing is the best on the evaluation of PGD-1 and PGD-2 examples. Based on the evaluation results from CIFAR10, we can see the differences between defensive methods are slightly larger. On all four kinds of white-box adversarial examples, Pure PGD is the best method and the test accuracy ranges from 48.33\% (PGD-1) to 56.18\% (FGSM). In the rest of defensive methods, GanDef is the best choice with test accuracy ranges from 45.62\% (PGD-1) to 54.14\% (FGSM).

Based on the comparison as well as visualization in Figure \ref{fig:visualization-TA}, it is clear that the proposed GanDef has the same level of performance as state-of-the-art adversarial training defenses in terms of the trained classifier's test accuracy on original and different adversarial examples.
\begin{table*}[tb]
    \begin{center}
    \begin{tabular}{ c  c | c  c  c  c  c  c }
    \hline 
    \multicolumn{2}{c}{} & Vanilla & Pure PGD & Mix PGD & Logit Pairing & GanDef & GanDef-Comb \\
    \hline
    \multirow{5}{*}{\begin{sideways}MNIST\end{sideways}} & Original & 98.70\% & 99.15\% & 99.17\% & 98.50\% & 99.10\% & 99.10\% \\
    & FGSM & 12.15\% & 97.60\% & 96.89\% & 97.00\% & 96.85\% & 96.85\% \\
    & BIM & 1.07\% & 94.75\% & 94.58\% & 95.83\% & 94.28\% & 94.28\% \\
    & PGD-1 & 0.87\% & 95.60\% & 95.56\% & 96.34\% & 95.31\% & 95.21\% \\
    & PGD-2 & 0.93\% & 94.14\% & 93.99\% & 95.42\% & 93.62\% & 93.38\% \\
    \hline
    \multirow{5}{*}{\begin{sideways}CIFAR10\end{sideways}} & Original & 89.69\% & 82.06\% & 83.70\% & 84.21\% & 84.05\% & 63.97\% \\
    & FGSM & 18.43\% & 56.18\% & 52.21\% & 51.63\% & 54.14\% & 87.61\% \\
    & BIM & 6.76\% & 49.21\% & 44.39\% & 44.09\% & 46.64\% & 76.02\% \\
    & PGD-1 & 6.48\% & 51.51\% & 47.11\% & 46.53\% & 49.21\% & 80.39\% \\
    & PGD-2 & 6.44\% & 48.33\% & 43.48\% & 43.28\% & 45.62\% & 73.56\% \\
    \hline
    \end{tabular}
    \end{center}
    \caption{Summary of Test Accuracy on Original and Adversarial Examples}
    \vspace{-4mm}
    \label{table:summary-TA}
\end{table*}
\begin{figure*}[tb]
\centering
\begin{minipage}[c]{.9\textwidth}
\centering
    \includegraphics[width=\linewidth]{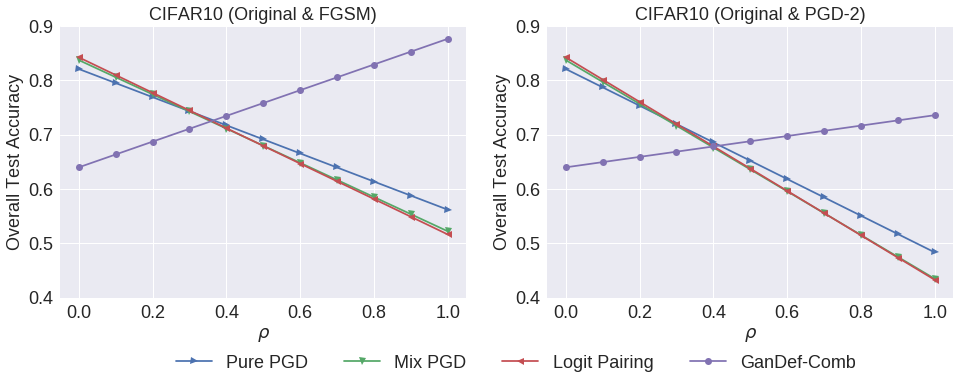}
\end{minipage}
\vspace{-4mm}
\caption{Visualization of Test Accuracy under Different Ratio of Adversarial Examples}
\label{fig:visualization-OA}
\end{figure*}
\subsection{Evaluation of GanDef-Comb}

In the second phase of evaluation, we consider GanDef-Comb which is a variant of GanDef. This variant utilizes both classifier and discriminator trained by GanDef. As we show in Section \ref{sec:defense}, the discriminator could be seen as a second line of defense when the trained classifier fails to make correct predictions on adversarial examples. By setting different threshold values for the discriminator, GanDef can dynamically control the trade-off between classifying original and adversarial examples. In current evaluation, the threshold is set to $0.5$.

On MNIST, the test accuracy of GanDef-Comb on original, FGSM and BIM examples is the same as that of GanDef. On PGD-1 and PGD-2 examples, the test accuracy of GanDef-Comb has a small degeneration (less than 0.3\%). This is because MNIST dataset is so simple such that the classifier alone can provide near optimal defense. Those misclassified corner cases are hard to be patched by utilizing discriminator. In common cases, the classifier has much larger degeneration on classifying adversarial examples. For example, on the CIFAR10, the benefit of utilizing discriminator is obvious due to such degeneration. From the results of test accuracy, GanDef-Comb is significantly better than state-of-the-art adversarial training defenses on mitigating FGSM, BIM, PGD-1 and PGD-2 examples. Based on the comparison, GanDef-Comb enhances test accuracy by at least 31.43\% on FGSM, 26.81\% on BIM, 28.88\% on PGD-1 and 25.23\% on PGD-2. Although the test accuracy of GanDef-Comb on original examples has about 20\% degeneration, the enhancement on defending adversarial examples benefits the overall test accuracy when the ratio of adversarial examples exceeds a certain limit.

To show the benefit of being able to control the trade-off, we design two experiments on CIFAR10 dataset. We form test dataset with original and adversarial examples (FGSM examples in the first experiment and PGD-2 examples in the second one). The ratio of adversarial examples, $\rho$, changes from $0$ to $1$. Giving similar weight losses in classifying original and adversarial examples, $\rho$ represents the probability of receiving adversarial examples. Or, giving similar probabilities of receiving original and adversarial examples, $\rho$ represents the weight of correctly classify adversarial examples ($1-\rho$ for original examples). These two evaluations are designed for risky or misbehavior-sensitive running environments, respectively.

The results of the overall test accuracy under different experiments are shown in Figure \ref{fig:visualization-OA}. It can be seen that GanDef-Comb is better than state-of-the-art defenses in terms of overall test accuracy when $\rho$ exceeds 41.7\%. In real applications, we could further enhance the overall test accuracy through changing the discriminator's threshold value. When $\rho$ is low, GanDef-Comb gives less attention to discriminator (high threshold value) and achieves similar performance as that of the state-of-the-art defenses. When $\rho$ is high, GanDef-Comb relies on discriminator (low threshold value) to detect more adversarial examples.

\section{Conclusion} \label{sec:conclusion}

In this paper, we introduce a new defensive method for Adversarial Examples, GanDef, which formulates a minimax game with a classifier and a discriminator during training. Through evaluation, we show that (1) the classifier achieves the same level of defense as classifiers trained by state-of-the-art defenses, and (2) using both classifier and discriminator (GanDef-Comb) can dynamically control the trade-off in classification and achieve higher overall test accuracy under the risky or misbehavior-sensitive running environment.

\section{Future Work} \label{sec:future}
One of the unsolved problem in the proposed GanDef method is that the degeneration of classifying original examples when the classifier and the discriminator are combined. For future work, we consider utilizing more sophisticated GAN models which can mitigate this degeneration.

%
%
%
\bibliographystyle{splncs04}
\bibliography{reference}

\begin{subappendices}
\renewcommand{\thesection}{}%

\section{Classifier Structures} \label{appendix:tables}

\begin{table}[H]
    \begin{center}
    \begin{tabular}{ c | c | c | c | c | c }
    \hline \hline
    Layer & Kernel Size & Strides & Padding & Activation & Init \\
    \hline \hline
    Convolution & $5 \times 5 \times 32$ & $1 \times 1$ & Same & ReLU & Default \\
    \hline
    MaxPool & $2 \times 2$ & $2 \times 2$ & - & - & - \\
    \hline
    Convolution & $5 \times 5 \times 64$ & $1 \times 1$ & Same & ReLU & Default \\
    \hline
    MaxPool & $2 \times 2$ & $2 \times 2$ & - & - & - \\
    \hline
    Flatten & - & - & - & - & - \\
    \hline
    Dense & $1024$ & - & - & ReLU & Default \\
    \hline
    Dense & $10$ & - & - & - & Default \\
    \end{tabular}
    \end{center}
    \caption{MNIST LeNet Classifier Structure}
    \label{table:mnist-classifier-structure}
    \vspace{-5mm}
\end{table}
\begin{table}[H]
    \begin{center}
    \begin{tabular}{ c | c | c | c | c | c }
    \hline \hline
    Layer & Kernel Size & Strides & Padding & Activation & Init \\
    \hline \hline
    Dropout & $0.2$ (drop rate) & - & - & - & - \\
    \hline
    Convolution & $3 \times 3 \times 96$ & $1 \times 1$ & Same & ReLU & He \\
    \hline
    Convolution & $3 \times 3 \times 96$ & $1 \times 1$ & Same & ReLU & He \\
    \hline
    Convolution & $3 \times 3 \times 96$ & $1 \times 1$ & Same & ReLU & He \\
    \hline
    MaxPool & $2 \times 2$ & $2 \times 2$ & - & - & - \\
    \hline
    Dropout & $0.5$ (drop rate) & - & - & - & - \\
    \hline
    Convolution & $3 \times 3 \times 192$ & $1 \times 1$ & Same & ReLU & He \\
    \hline
    Convolution & $3 \times 3 \times 192$ & $1 \times 1$ & Same & ReLU & He \\
    \hline
    Convolution & $3 \times 3 \times 192$ & $1 \times 1$ & Same & ReLU & He \\
    \hline
    MaxPool & $2 \times 2$ & $2 \times 2$ & - & - & - \\
    \hline
    Dropout & $0.5$ (drop rate) & - & - & - & - \\
    \hline
    Convolution & $3 \times 3 \times 192$ & $1 \times 1$ & Valid & ReLU & He \\
    \hline
    Convolution & $1 \times 1 \times 192$ & $1 \times 1$ & Same & ReLU & He \\
    \hline
    Convolution & $1 \times 1 \times 192$ & $1 \times 1$ & Same & ReLU & He \\
    \hline
    GlobalAvgPool & - & - & - & - & - \\
    \hline
    Dense & $10$ & - & - & - & Default \\
    \end{tabular}
    \end{center}
    \caption{CIFAR10 allCNN Classifier Structure}
    \label{table:cifar-classifier2-structure}
    \vspace{-5mm}
\end{table}

\end{subappendices}
\end{document}